\documentclass{melba}
\usepackage{mwe} % to get dummy images, only for the example
\usepackage{amsmath,amsfonts}
\melbaid{YYYY:NNN}  % This is provided upon by the publishing editor
\doi{10.59275/j.melba.2024-AAAA}
\melbaauthors{Kondrateva, Barg and Vasiliev}  % Note: this one is also used to set the pdf 'authors' metadata
\email{ekaterina.kondrateva@maastrichtuniversity.nl}
\volume{2}
\firstpageno{1337}  % Communicated by the publishing editor
\melbayear{2025}  % The publication year
\datesubmitted{yyyy-m1-d1}  % Date submitted to MELBA: mm/yyyy
\datepublished{yyyy-m2-d2}  % Today's date: mm/yyyy

\melbaspecialissue{Medical Imaging with Deep Learning (MIDL) 2025}
\melbaspecialissueeditors{Marleen de Bruijne, Tal Arbel, Ismail Ben Ayed, Hervé Lombaert}

\ShortHeadings{Brain MRI Segmentation Across Scanners}{Kondrateva, Barg and Vasiliev}

\title{Benchmarking the Reproducibility of Brain MRI Segmentation Across Scanners and Time}

\author{
	\firstname Ekaterina \surname Kondrateva\orcid{0000-0003-3623-6106}
	\firstname  Sandzhi  \surname Barg\orcid{0009-0002-6498-0282}
        \firstname Mikhail \surname Vasiliev
}
\affiliations{% <- trailing '%' to avoid unwanted indent
}

\abstract{%   <- trailing '%' for backward compatibility of .sty file
Accurate and reproducible brain morphometry from structural MRI is critical for monitoring neuroanatomical changes across time and imaging domains. Although deep learning has accelerated segmentation workflows, scanner-induced variability and reproducibility limitations remain—particularly in longitudinal and multi-site settings. In this study, we benchmark two state-of-the-art pipelines—\textit{FastSurfer} and \textit{SynthSeg}—both integrated into \textit{FreeSurfer}, one of the most widely adopted tools in neuroimaging. Using two complementary datasets—a 17-year single-subject longitudinal cohort (SIMON) and a 9-site test-retest cohort (SRPBS)—we quantify inter-scan segmentation variability using Dice, Surface Dice, Hausdorff Distance (HD95), and Mean Absolute Percentage Error (MAPE).
Our results reveal up to 7–8\% volume variation in small subcortical structures such as the amygdala and ventral diencephalon, even under controlled test-retest conditions. This raises a critical question: is it feasible to detect subtle longitudinal changes—on the order of 5–10\%—in pea-sized brain regions, given the magnitude of domain-induced morphometric noise? We further analyze the effects of registration choices and interpolation modes, and propose surface-based quality filtering to improve reliability. This work provides a reproducible benchmark and calls for harmonization strategies to enable robust morphometry in real-world neuroimaging studies.
	Our code is available at~\url{https://github.com/kondratevakate/brain-mri-segmentation}.}


\begin{document}

% top matter
\twocolumn[\maketitle]
% comment the preceedings and uncomment the following if the authors list + abstract is longer than one page
% \maketitle
% \twocolumn

% Introduction (or first section)
% \rule{\textwidth}{1pt}
\section{Introduction}
   \enluminure{A}{dvances} in AI-driven medical imaging have revolutionized pathology detection, yet reproducible morphometric analysis of healthy brains—especially across scanners and over time—remains a challenge. This gap limits our ability to monitor individual brain health trajectories and detect early pathological changes. While artificial intelligence (AI) has significantly advanced medical imaging—particularly in pathology segmentation tasks such as tumor identification in the BraTS challenge—there remains a notable gap in applying these advancements to morphometric analyses of healthy brains across varied domains. This underexplored area presents opportunities for developing robust, generalizable AI models that can accurately capture subtle anatomical variations, thereby deepening insight into brain aging and development.
    
    Traditional tools like FreeSurfer~\citep{fischl2012freesurfer} have been instrumental in providing detailed morphometric analyses. Recent integrations, such as SynthSeg~\citep{billot2023synthseg}, offer contrast-agnostic segmentation capabilities trained on synthetic data, aiming to improve generalizability across different imaging protocols. Despite these advancements, challenges persist in ensuring reproducibility of volumetric estimates under real-world conditions, particularly when dealing with data from multiple scanners and protocols.
    
    This study aims to assess the consistency of brain volume measurements using FastSurefer and FreeSurfer 8 with integrated SynthSeg across longitudinal MRI scans from a single individual. By quantifying inter-scan variability using metrics like absolute volume difference, Dice, and Surface Dice, we seek to highlight the limitations of current segmentation pipelines in personalized brain health monitoring and early detection of neurodegenerative conditions.

%%%%%%%%%%%%%%%%%%%%%%%%%%%%%%%%%%%%%%%%%%%%%%%%%%%%%%%%%%%%%%%%%%%%%%%%%%%
% Related works
%%%%%%%%%%%%%%%%%%%%%%%%%%%%%%%%%%%%%%%%%%%%%%%%%%%%%%%%%%%%%%%%%%%%%%%%%%%
% Make sure to put your work into context and include apporpriate citations.
% We do not have limits on citation counts.
\section{Related Works}
	
    \section*{Related Work}

    \subsection{Segmentation Pipelines for Morphometry Extraction}
    
    Deep learning has significantly advanced individual-level brain morphometry from structural MRI. Traditional pipelines such as \textit{FreeSurfer}~\citep{fischl2012freesurfer} have long served as a gold standard, producing cortical and subcortical morphometric features (e.g., thickness, volume, surface area). However, these methods are computationally intensive and sensitive to scanner variability, limiting their scalability in large-scale or multisite studies.
    
    Recent versions of FreeSurfer integrate \textit{SynthSeg}~\citep{billot2023synthseg}, a contrast-agnostic segmentation model trained on synthetic data. \textit{SynthSeg+} provides robust volumetric estimates across diverse contrasts, resolutions, and scanners. Its compatibility with standard atlases (e.g., Desikan-Killiany, MUSE) makes it suitable for harmonized morphometry across heterogeneous datasets.
    
    To address runtime bottlenecks, \textit{FastSurferVINN}~\citep{henschel2023fastsurfervinn} replaces FreeSurfer’s anatomical stream with a vision transformer-based model, enabling accurate surface-based cortical thickness estimation within minutes. Tools such as Brainchop prioritize clinical scalability, though often at the cost of generalization to unseen protocols.
    
    Other high-performing segmentation models include \textit{nnU-Net}~\citep{isensee2021nnu} and \textit{nnFormer}~(\citep{zhou2023nnformer}), which yield excellent accuracy in controlled benchmarks but often require dataset-specific finetuning to generalize effectively in clinical or real-world settings.
    
    Recent segmentation advances also include multi-atlas deep learning pipelines~\citep{wang2025multiatlas}, which integrate lifespan-spanning templates to enhance anatomical precision, particularly in pediatric and geriatric cohorts.
    
    \subsection{Longitudinal Modeling and Individualized Morphometry}
    
    Beyond segmentation, recent work has focused on modeling spatiotemporal brain changes at the individual level. \textit{Latent diffusion-based progression modeling}, such as Brain Latent Progression (BLP)~\citep{puglisi2025blp}, uses temporally conditioned diffusion models to infer personalized disease trajectories from serial MRI scans.
    
    \textit{Learning-based Inference of Brain Change (LIBC)}~\citep{kim2025libc} models smooth morphometric changes over time using neural timeline embeddings, capturing subtle age- and disease-related progression in cortical and subcortical structures.
    
    Normative modeling frameworks \citep{allen2024normative} enable the estimation of z-score deviations from large-scale population references. This approach is particularly effective in identifying early deviations in psychiatric populations and supports both clinical and subclinical applications.
    
    Another widely adopted line of work focuses on brain age prediction. \textit{BrainAGE}~\citep{franke2012brainage} models estimate biological aging based on MRI-derived morphometric features, frequently using \textit{FreeSurfer} outputs. These models have demonstrated strong longitudinal reliability and clinical interpretability.
    
    Emerging tools like \textit{Neurofind}~\citep{neurofind2025} offer user-friendly platforms that integrate normative modeling and brain age estimation, providing individualized reports based on high-resolution structural MRI images.
    
    Despite these advances, challenges remain in achieving sulcal-level surface precision, quantifying uncertainty, and ensuring reproducibility in real-world multisite studies. Although morphometry has clear clinical applications (e.g., in epilepsy and dementia\footnote{\url{https://icometrix.com/expertise\#mri}}), rigorous longitudinal reproducibility benchmarks remain scarce.

    \subsection{Brain morpometry as a biomarker}

    Longitudinal MRI studies have greatly expanded our understanding of how brain morphometry changes over time, particularly in response to aging, disease, and stress. A growing body of work highlights structural biomarkers in specific brain regions—especially the hippocampus, anterior cingulate, and prefrontal cortex—that reflect vulnerability or resilience to neuropsychiatric conditions.

    In healthy populations \cite{papagni2011stress} demonstrated gray matter volume (GMV) reductions in the anterior cingulate cortex (ACC), hippocampus, and medial prefrontal cortex (mPFC) in individuals exposed to stress. Similar findings were confirmed in large-scale aging studies, including \cite{schaefer2018longitudinal}, who reported consistent hippocampal atrophy associated with aging. \cite{macdonald2021mri} provide a broader review of region-specific atrophy across the lifespan. Structural biomarkers also inform psychiatric research. \cite{cardoner2024impact} review evidence of stress-induced degeneration in the ACC and dorsolateral prefrontal cortex (dlPFC), while \cite{carnevali2018resilience} identify preserved amygdala volumes as potential resilience markers. UK Biobank analyses further support longitudinal volume reductions in fronto-limbic circuits among individuals with high stress exposure~\citep{statsenko2022brain}. Importantly, several studies have examined structural changes within individuals undergoing therapy. Gryglewski et al.~\cite{gryglewski2019ect} found hippocampal and amygdalar volume increases after electroconvulsive therapy (ECT) in treatment-resistant depression. \cite{furtado2012tms} reported volumetric growth in the dlPFC after rTMS. \cite{frodl2008gmd} showed that psychotherapy attenuated gray matter loss over three years in depression.  Together, these findings suggest that MRI-based brain morphometry, especially when assessed longitudinally, provides meaningful biomarkers for brain health across both normative and pathological aging.

% A methodological, model, or similar section often comes here.
\section{Methods}

        We study reproducibility of brain MRI segmentation pipelines across longitudinal and multi-site datasets. We use two publicly available datasets—SIMON and SRPBS—spanning a wide range of scanners and protocols. We compare segmentation outputs from FreeSurfer~8.0.0, FastSurfer, and SynthSeg, using FreeSurfer's \texttt{recon-all} pipeline as a reference. Segmentation reproducibility is evaluated using a targeted subset of cortical and subcortical ROIs most relevant for neuroimaging biomarkers. For surface-based comparisons, we apply rigid registration using ANTs and assess the effect of different interpolation modes and reference spaces. Quantitative evaluation is performed using Dice coefficient, Surface Dice, 95th percentile Hausdorff distance (HD95), and mean absolute percentage error (MAPE) of regional brain volumes. 
    
        \subsection{Data}

        We utilized two datasets for our analysis:
        
        \textbf{SIMON Dataset:} This dataset comprises 73 T1-weighted MRI scans of a single healthy male subject, collected over 17 years across multiple sites and 1.5T scanners~\citep{duchesne2019canadian}.
        
        \textbf{SRPBS Traveling Subject Dataset:} This dataset includes 411 T1-weighted MRI scans from 9 healthy subjects, each scanned at 9 different sites using 3T MRI scanners. The data is organized following the BIDS format and includes accompanying metadata such as participant demographics and scanner parameters~\citep{tanaka2021multi}. A detailed comparison of acquisition parameters between the SIMON and SRPBS datasets is provided in Table~\ref{tab:acquisition_parameters}.

       \begin{table}[ht]  
        \centering  
        \caption{Acquisition parameters}  
        \begin{tabular}{@{}lccc@{}}  
        \toprule  
        \textbf{Acquisition parameter} & \textbf{SIMON} & \textbf{SRPBS\_TS} \\ \midrule  
        \textbf{Age}                   &        \\   
        min                            & 29     & 24 \\   
        max                            & 46     & 32 \\   
        \#unique                       & 1      & 9 \\ \midrule
        \textbf{Test-retest time, days} &        \\   
        min                            & 0      & 1 \\   
        max                            & 1154   &  \\   
        \#unique                       & 45     & 143 \\ \midrule
        \textbf{Echo Time, ms}         &        \\   
        min                            & 0.002  & 0.001 \\   
        max                            & 0.003  & 0.003 \\   
        \#unique                       & 8      & 24 \\ \midrule  
        \textbf{Repetition Time, ms}   &        \\   
        min                            & 0.007  & 0.007 \\   
        max                            & 2.3    & 2.3 \\   
        \#unique                       & 8      & 26 \\ \midrule  
        \textbf{Voxel volume, x}       &        \\   
        min                            & 0.8    & 0.8 \\   
        max                            & 1.1    & 1.2 \\   
        \#unique                       & 6      & 35 \\ \midrule
        \textbf{Voxel volume, y}       &        \\   
        min                            & 0.8    & 0.7 \\   
        max                            & 1.0    & 1.0 \\   
        \#unique                       & 4      & 8 \\ \midrule
        \textbf{Voxel volume, z}       &        \\   
        min                            & 0.8    & 0.7 \\   
        max                            & 1.0    & 1.0 \\   
        \#unique                       & 4      & 14 \\ \bottomrule  
        \end{tabular}  
        \label{tab:acquisition_parameters}  
        \end{table}  

	\subsection{Segmentation} 

        We employed FreeSurfer~8.0.0 (released February 27, 2025) for cortical surface reconstruction and anatomical segmentation using the \texttt{recon-all} pipeline. To evaluate segmentation performance, we compared two state-of-the-art deep learning-based methods: FastSurfer~\cite{henschel2020fastsurfer} and SynthSeg~\cite{billot2023synthseg}. FastSurfer offers rapid and accurate whole-brain segmentation, replicating FreeSurfer's anatomical outputs, while SynthSeg provides robust segmentation across varying MRI contrasts and resolutions without the need for retraining. For consistency and comprehensive analysis, we selected FreeSurfer's \texttt{recon-all} outputs as the reference standard and assessed the Desikan-Killiany-Tourville (DKT) atlas parcellations, encompassing 100 cortical and subcortical regions.

        \subsection{Registration}
        For surface-based metrics, we applied rigid-body registration using ANTs~\cite{Avants2011}, computing transforms from the original T1-weighted images. We evaluated two interpolation mode \texttt{nearestNeighbor}. Registrations were performed either to the subject's first session or to an asymmetric MRI atlas. This approach aimed to assess the impact of interpolation schemes and reference spaces on the consistency of surface-derived measurements.
        
        \subsection{ROI Analysis}
        
        We focused our analysis on 9 cortical and 8 subcortical bilateral regions of interest (ROIs), selected based on their relevance as biomarkers in neuroimaging studies. The complete list of analyzed ROIs is provided in Table~\ref{tab:roi_list}. Differences observed across successive MRI sessions were interpreted as domain variations.

\subsection{Metrics}
To evaluate segmentation reproducibility, we report absolute volume differences, as well as spatial similarity metrics: Dice coefficient, Surface Dice, and 95th percentile Hausdorff Distance (HD95). Each metric captures a different aspect of agreement between two segmentations: volumetric overlap, boundary proximity, and outlier misalignment. These are computed for each region of interest (ROI) and aggregated across sessions.

\paragraph{Dice Coefficient (DSC):}
Dice measures the voxel-level overlap between two binary masks $A$ and $B$ (e.g., predicted and reference segmentations):
\begin{equation}
\mathrm{DSC} = \frac{2|A \cap B|}{|A| + |B|}
\label{eq:dice}
\end{equation}
Here, $|A|$ and $|B|$ are the number of voxels in each mask, and $|A \cap B|$ is the number of voxels they share. Dice is widely used due to its simplicity, but can be insensitive to boundary errors.

\paragraph{Surface Dice (S-DSC):}
Surface Dice quantifies the proportion of surface points that lie within a distance $\tau$ between the two segmentation boundaries $\partial A$ and $\partial B$:
\begin{align}
\mathrm{S\text{-}DSC} = 
& \frac{|\{x \in \partial A : d(x, \partial B) \leq \tau\}|}{|\partial A| + |\partial B|} \notag \\
+\, & \frac{|\{y \in \partial B : d(y, \partial A) \leq \tau\}|}{|\partial A| + |\partial B|}
\label{eq:surfacedice}
\end{align}
Here, $d(x, \partial B)$ denotes the minimum Euclidean distance from a point $x$ on the surface of $A$ to the surface of $B$, and $\tau$ is the distance tolerance (set to 1\,mm in our experiments). This metric captures small surface deviations and is well-suited for assessing perceptual segmentation accuracy.

\paragraph{95th Percentile Hausdorff Distance (HD95):}
HD95 captures the worst-case boundary discrepancy, ignoring extreme outliers by focusing on the 95th percentile of all boundary distances:
\begin{multline}
\mathrm{HD}_{95}(A, B) = \max\bigg\{ \mathrm{P}_{95}\left(\{d(x, \partial B) : x \in \partial A\}\right), \\
\mathrm{P}_{95}\left(\{d(y, \partial A) : y \in \partial B\}\right) \bigg\}
\label{eq:hd95}
\end{multline}
Where $\mathrm{P}_{95}$ denotes the 95th percentile, and $d(x, \partial B)$ is the shortest distance from point $x$ to the other surface. HD95 is useful for identifying large local deviations in shape or topology.

\paragraph{Mean Absolute Percentage Error (MAPE):}
To compare volumes across repeated scans, we use the mean absolute percentage error between segmentation volumes:
\begin{equation}
\mathrm{MAPE} = \frac{100\%}{n} \sum_{i=1}^n \left| \frac{V_i^{\text{pred}} - V_i^{\text{ref}}}{V_i^{\text{ref}}} \right|
\label{eq:mape}
\end{equation}
Where $V_i^{\text{pred}}$ and $V_i^{\text{ref}}$ are the predicted and reference volumes for region $i$, and $n$ is the number of ROIs. MAPE is intuitive for assessing how much segmentations deviate from expected anatomical volumes.

        \subsection{Computations}
        All experiments were conducted on a Google Cloud Platform (GCP) instance equipped with 64~vCPUs and 512~GB of RAM. FreeSurfer~8.0.0 was executed using a single CPU core per subject, with an average processing time of approximately 2 hours per subject. Attempts to utilize GPU acceleration for SynthSeg were unsuccessful due to driver compatibility issues, resulting in all SynthSeg processing being performed on the CPU.

\section{Results}

        \subsection{SRPBS Test-Retest: FastSurfer}
        
\begin{figure*}[ht]
        \centering
        \includegraphics[width=\linewidth]{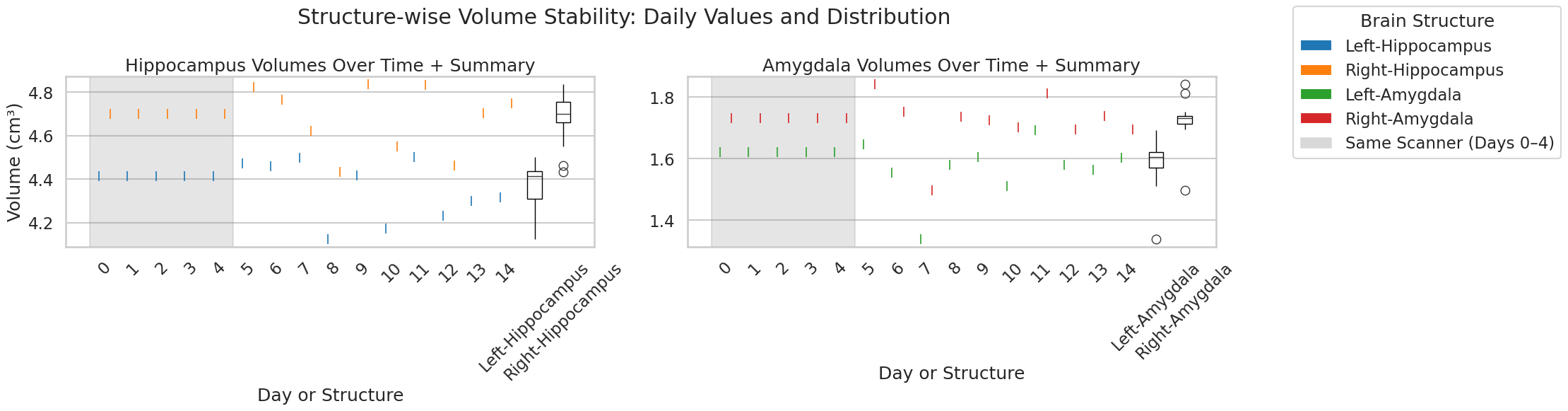}
        \caption{Volume stability for left/right hippocampus and amygdala across Subject $1$, 15 sessions in SRPBS Traveling Subject dataset. FastSurfer results with ANTS registration. The first 5 days (shaded) were acquired on the same scanner; subsequent sessions were acquired at different sites.}
        \label{fig:srpbs_fastsurfer_stability}
        \end{figure*}

        We analyzed 15 sessions from the SRPBS Traveling Subject dataset~\citep{tanaka2021multi} using FastSurfer . As shown in Figure~\ref{fig:srpbs_fastsurfer_stability}, the first five sessions were acquired on the same scanner across five consecutive days, while the remaining sessions involved different scanners and sites.
        
        For both hippocampus and amygdala, volume estimates during the same-scanner phase were highly consistent. For example, left hippocampus volumes ranged narrowly between 4.42–4.44\,cm³ (SD = 0.01), and right amygdala volumes ranged from 1.73–1.75\,cm³ (SD = 0.008). In contrast, sessions from different scanners showed noticeable variability: left hippocampus ranged from 4.16–4.53\,cm³ (SD = 0.10), and right amygdala from 1.50–1.85\,cm³ (SD = 0.11).
        
        This highlights that even in a highly controlled test-retest design, inter-scanner variability introduces morphometric noise of up to 10\%, especially in small structures like the amygdala. Reliable quantification in longitudinal or multisite settings requires either harmonization or robust outlier filtering.

		\subsection{SIMON Longitudinal: FastSurfer vs. SynthSeg}

We evaluated segmentation reproducibility across 73 sessions over 17 years using FastSurfer and SynthSeg.

\begin{figure}[ht]
    \centering
    \includegraphics[width=\columnwidth]{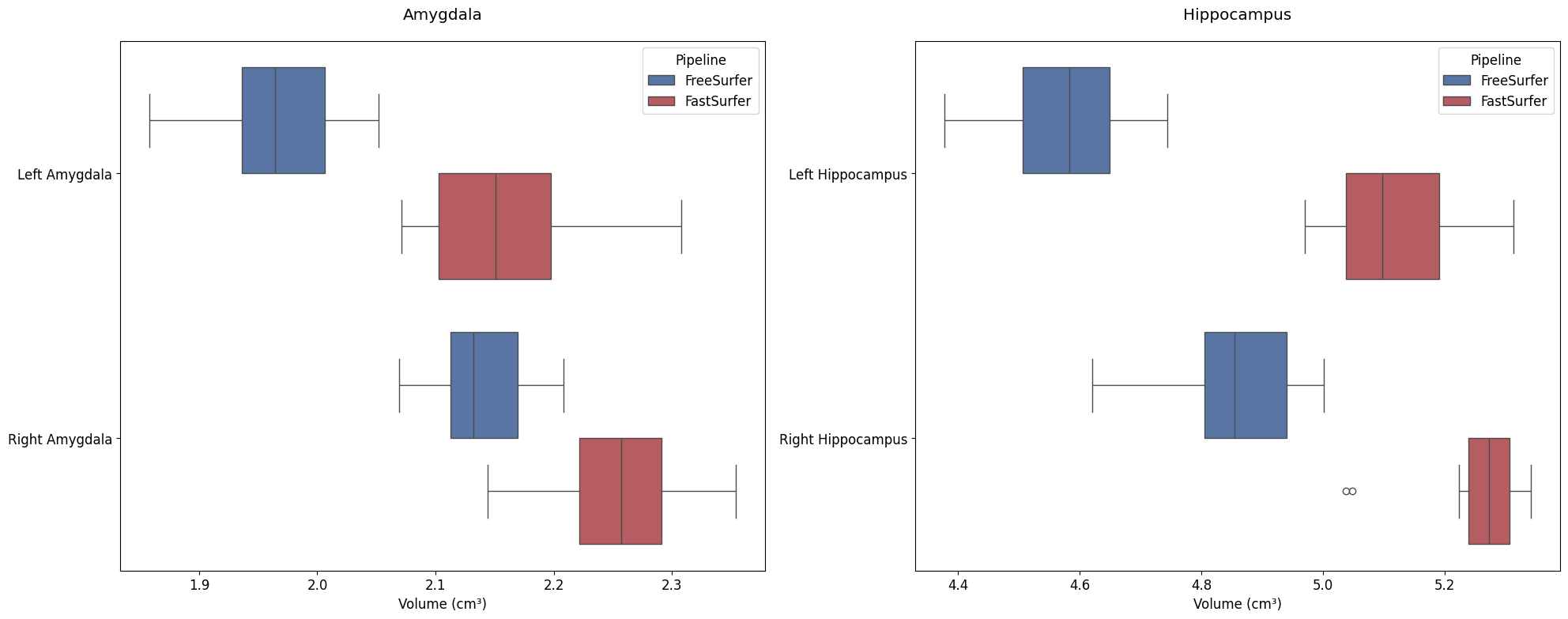}
    \caption{SIMON dataset: Comparison of volume distributions from FastSurfer and SynthSeg for Amygdala and Hippocampus, y-axis denotes volume in cm³.}
    \label{fig:simon_comparison_boxplot}
\end{figure}

\paragraph{FastSurfer.}
FastSurfer \texttt{recon-all} failed on 3 sessions and 8 runs. For valid outputs, subcortical volumes were stable:  
Left/Right Amygdala: $1.93 \pm 0.17$ / $2.10 \pm 0.12$~cm³  
Left/Right Hippocampus: $4.54 \pm 0.19$ / $4.82 \pm 0.16$~cm³  
Volume trajectories showed small upward trends ($R^2$ = 0.12–0.26).\ref{fig:simon_comparison_boxplot}

\begin{figure}[ht]
    \centering
    \includegraphics[width=\columnwidth]{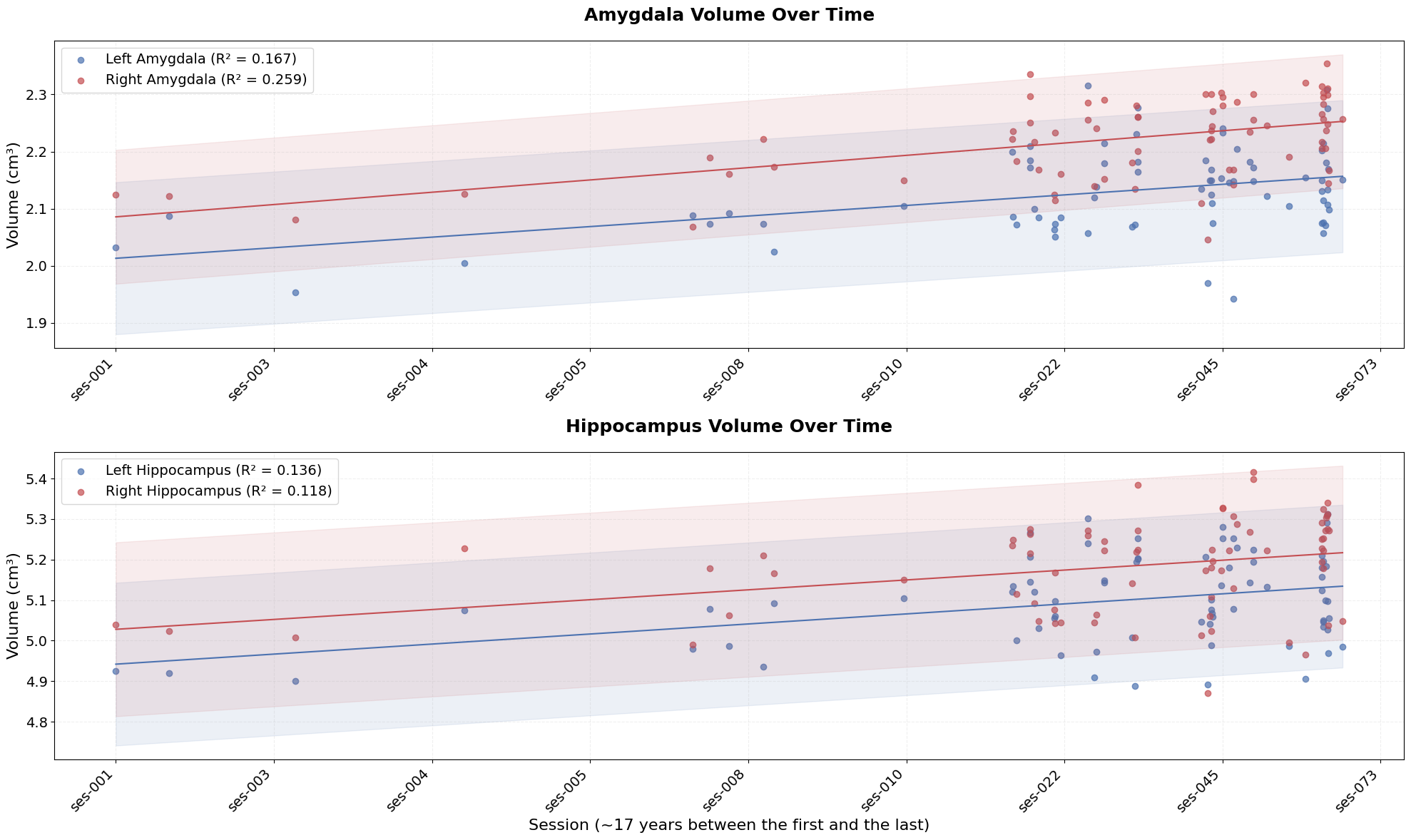}
    \caption{SIMON dataset: Volume trajectories of Amygdala and Hippocampus over time for 73 MRI scans in 17 years for one healthy individual using SynthSeg. Confidence intervals and regression trends are shown.}
    \label{fig:synthseg_regression}
\end{figure}

\paragraph{SynthSeg.}
Subcortical variation averaged 3.1\%, peaking at 15–20\%.  
Cortical parcellations varied by 5\% on average, with outliers exceeding 40–90\%.  
Volumes were consistently higher:  
Amygdala: $2.13 \pm 0.07$ / $2.22 \pm 0.07$~cm³  
Hippocampus: $5.10 \pm 0.11$ / $5.18 \pm 0.12$~cm³  
(Figure~\ref{fig:synthseg_regression})

Volume comparisons show that FastSurfer consistently estimates larger volumes than SynthSeg. For example, the left hippocampus volume averaged $5.12 \pm 0.12\,\text{cm}^3$ in FastSurfer versus $4.58 \pm 0.12\,\text{cm}^3$ in SynthSeg.

Table~\ref{tab:segmentation_subcortical_full} compares FreeSurfer and FastSurfer across eight representative cortical structures. FastSurfer yielded consistently higher Dice scores (e.g., 0.861 vs. 0.793 for Insula, 0.816 vs. 0.728 for Fusiform), suggesting improved anatomical overlap. Surface Dice values remained comparable, with minimal variation between methods. Volume differences were notably smaller in FastSurfer (e.g., 2.0\,mm³ for Insula, compared to 31.6\,mm³ in FreeSurfer), reflecting reduced bias. Interestingly, FreeSurfer produced lower Hausdorff distances in some regions (e.g., Superior Frontal Cortex: 1.21\,mm vs. 1.74\,mm), but at the cost of greater volume deviation. Overall, FastSurfer offers more consistent cortical segmentation while maintaining competitive boundary accuracy.

\begin{table*}[t]
\centering
\caption{Comparison of SynthSeg in FreeSurfer 8 (FS) and FastSurfer (Fast) segmentation performance across subcortical structures. Volume differences are in mm\textsuperscript{3}, Dice and Surface Dice are unitless, HD95 is in mm.}
\label{tab:segmentation_subcortical_full}
\small
\resizebox{\textwidth}{!}{%
\begin{tabular}{l|cc|cc|cc|cc|cc|cc|cc|cc}
\toprule
\textbf{Metric} & \multicolumn{2}{c|}{\textbf{Accumbens}} & \multicolumn{2}{c|}{\textbf{Amygdala}} & \multicolumn{2}{c|}{\textbf{Caudate}} & \multicolumn{2}{c|}{\textbf{Hippocampus}} & \multicolumn{2}{c|}{\textbf{Pallidum}} & \multicolumn{2}{c|}{\textbf{Putamen}} & \multicolumn{2}{c|}{\textbf{Thalamus}} & \multicolumn{2}{c}{\textbf{Ventral DC}} \\
 & FS & Fast & FS & Fast & FS & Fast & FS & Fast & FS & Fast & FS & Fast & FS & Fast & FS & Fast \\
\midrule
\textbf{Volume Diff (mm\textsuperscript{3})} & 5.20 & -0.56 & 0.22 & -2.23 & 14.18 & 1.36 & 12.46 & -0.17 & 11.99 & 1.94 & 19.99 & -5.04 & 2.27 & 8.30 & 12.32 & 1.06 \\
\textbf{Dice}               & 0.803 & 0.827 & 0.858 & 0.862 & 0.868 & 0.874 & 0.850 & 0.868 & 0.850 & 0.859 & 0.897 & 0.902 & 0.909 & 0.917 & 0.858 & 0.873 \\
\textbf{Surface Dice}       & 0.965 & 0.955 & 0.961 & 0.944 & 0.972 & 0.957 & 0.964 & 0.963 & 0.958 & 0.927 & 0.969 & 0.956 & 0.947 & 0.948 & 0.959 & 0.950 \\
\textbf{HD95 (mm)}          & 1.23  & 1.60  & 1.26  & 1.50  & 1.20  & 1.56  & 1.23  & 1.34  & 1.27  & 1.64  & 1.21  & 1.58  & 1.33  & 1.45  & 1.23  & 1.43 \\
\bottomrule
\end{tabular}
}
\end{table*}

\begin{table*}[t]
\centering
\caption{Comparison of SynthSeg in FreeSurfer 8 (FS) and FastSurfer segmentation performance across selected cortical structures. Volume difference is in mm\textsuperscript{3}, Dice and Surface Dice are unitless, HD95 is in mm.}
\label{tab:segmentation_cortical_fs_fast}
\scriptsize
\resizebox{\textwidth}{!}{%
\begin{tabular}{l|cc|cc|cc|cc|cc|cc|cc|cc|cc}
\toprule
\textbf{Metric} 
& \multicolumn{2}{c|}{\textbf{Caudal Ant. Cingulate}} 
& \multicolumn{2}{c|}{\textbf{Entorhinal Cortex}} 
& \multicolumn{2}{c|}{\textbf{Fusiform Gyrus}} 
& \multicolumn{2}{c|}{\textbf{Inferior Parietal}} 
& \multicolumn{2}{c|}{\textbf{Insula}} 
& \multicolumn{2}{c|}{\textbf{Lat. Orbitofrontal}} 
& \multicolumn{2}{c|}{\textbf{Med. Orbitofrontal}} 
& \multicolumn{2}{c|}{\textbf{Superior Frontal}} 
& \multicolumn{2}{c}{\textbf{Superior Temporal}} \\
\cmidrule{2-19}
& FS & Fast & FS & Fast & FS & Fast & FS & Fast & FS & Fast & FS & Fast & FS & Fast & FS & Fast & FS & Fast \\
\midrule
\textbf{Volume Diff}     
& 21.62 & 2.90  & 7.75 & -4.78 & 58.67 & -2.95 & 113.35 & 3.51 
& 31.60 & 2.00  & 64.48 & 7.46  & 35.29 & 5.04  & 216.32 & 42.99 & 115.34 & 15.80 \\
\textbf{Dice}            
& 0.746 & 0.820 & 0.709 & 0.794 & 0.728 & 0.816 & 0.726 & 0.807 
& 0.793 & 0.861 & 0.712 & 0.796 & 0.663 & 0.780 & 0.733 & 0.807 & 0.759 & 0.817 \\
\textbf{Surface Dice}    
& 0.965 & 0.958 & 0.922 & 0.922 & 0.959 & 0.964 & 0.973 & 0.963 
& 0.970 & 0.971 & 0.952 & 0.948 & 0.938 & 0.949 & 0.970 & 0.966 & 0.964 & 0.958 \\
\textbf{HD95}            
& 1.24 & 1.64  & 1.72 & 1.72  & 1.28 & 1.35  & 1.19 & 1.47 
& 1.35 & 1.51  & 1.34 & 1.85  & 1.46 & 1.84  & 1.21 & 1.74  & 1.26 & 1.47 \\
\bottomrule
\end{tabular}
}
\end{table*}

\begin{figure*}[h]
        		\centering
                \includegraphics[width=\textwidth]{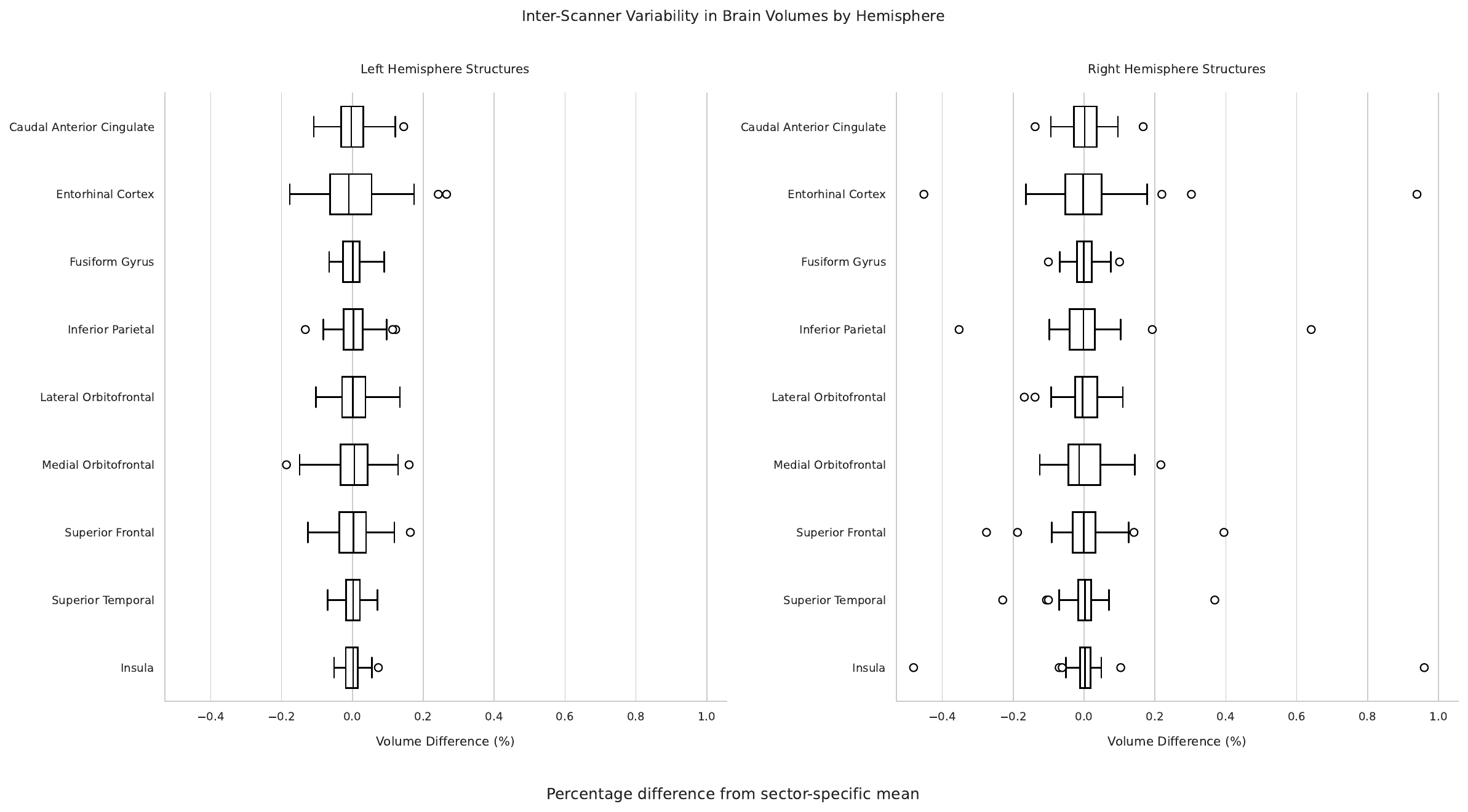}
        		\caption{Inter-scanner variability of cortical volumes in the SIMON dataset. Boxplots show DICE and Surface DICE metrics between consecutive scans, grouped by hemisphere. }
        	\end{figure*}

            \begin{figure*}[h]
    		\centering
            \includegraphics[width=\textwidth]{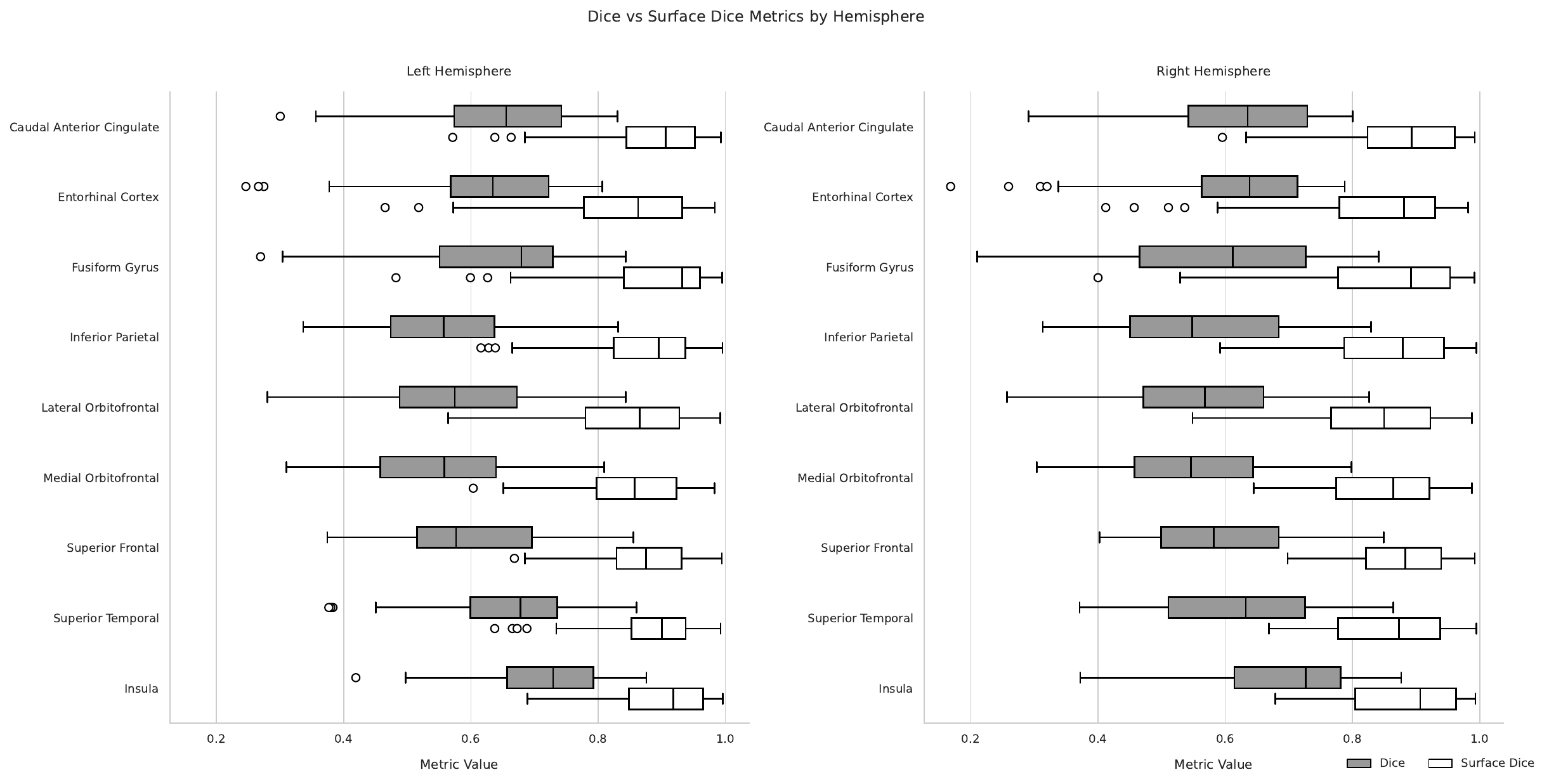}
    		\caption{Inter-scanner variability of cortical volumes in the SIMON dataset. Boxplots show the percentage difference from the structure-specific mean across repeated sessions, grouped by hemisphere. }

    	\end{figure*}

            \begin{figure*}[h]
        		\centering
                \includegraphics[width=\textwidth]{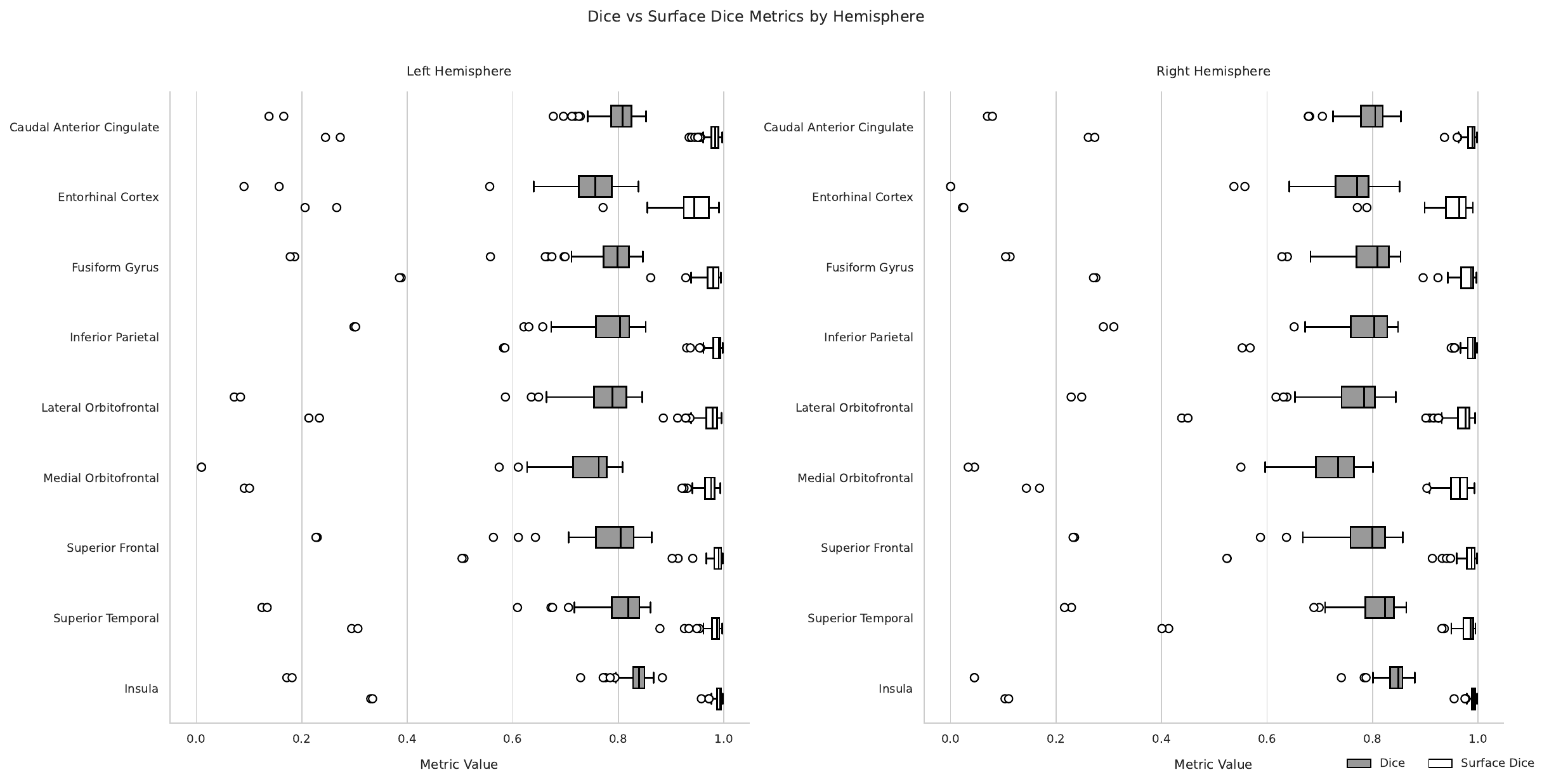}
        		\caption{Dice and Surface Dice coefficient distributions across cortical regions in the left and right hemispheres of the SIMON dataset using SynthSeg. Each structure is evaluated over multiple longitudinal scans from the same individual. Surface Dice (white boxes) consistently exceeds traditional Dice (gray boxes), especially in regions with complex geometry such as the entorhinal cortex and insula.}

        	\end{figure*}

\begin{table*}[htbp]  
\centering  
\resizebox{\textwidth}{!}{%  
\begin{tabular}{l|cc|cc|cc|cc|cc|cc|cc|cc}  
\hline  
\multirow{2}{*}{Metric} & \multicolumn{2}{c|}{Accumbens} & \multicolumn{2}{c|}{Amygdala} & \multicolumn{2}{c|}{Caudate} & \multicolumn{2}{c|}{Hippocampus} & \multicolumn{2}{c|}{Pallidum} & \multicolumn{2}{c|}{Putamen} & \multicolumn{2}{c|}{Thalamus} & \multicolumn{2}{c}{Ventral DC} \\
 & SRPBS & SIMON & SRPBS & SIMON & SRPBS & SIMON & SRPBS & SIMON & SRPBS & SIMON & SRPBS & SIMON & SRPBS & SIMON & SRPBS & SIMON \\
\hline  
Volume diff (cm³)   & 0.046 & 0.030 & 0.102 & 0.076 & 0.119 & 0.098 & 0.207 & 0.125 & 0.102 & 0.095 & 0.206 & 0.136 & 0.450 & 0.374 & 0.219 & 0.141 \\
Dice                & 0.677 & 0.803 & 0.790 & 0.858 & 0.802 & 0.868 & 0.782 & 0.850 & 0.789 & 0.850 & 0.848 & 0.897 & 0.868 & 0.909 & 0.806 & 0.858 \\
Surface Dice        & 0.849 & 0.965 & 0.840 & 0.961 & 0.868 & 0.972 & 0.845 & 0.964 & 0.843 & 0.958 & 0.870 & 0.969 & 0.820 & 0.947 & 0.873 & 0.959 \\
HD95 (mm)           & 1.735 & 1.228 & 1.697 & 1.263 & 1.584 & 1.200 & 1.830 & 1.234 & 1.675 & 1.271 & 1.582 & 1.210 & 1.828 & 1.327 & 1.620 & 1.233 \\
\hline  
\end{tabular}%  
}  
\caption{Comparison of segmentation metrics between the SRPBS and SIMON datasets across subcortical structures. Volume difference is shown in $\text{cm}^3$, Dice and Surface Dice are unitless similarity scores, and HD95 represents the 95th percentile Hausdorff distance in millimeters.}  
\label{tab:subcortical_comparison}  
\end{table*}

\subsection{Comparison of Distance Metrics Across Datasets}

Table~\ref{tab:subcortical_comparison} summarizes segmentation reproducibility across eight subcortical structures in the SRPBS and SIMON datasets. Volume differences (in cm³) were consistently higher in SRPBS, reflecting greater domain variability due to inter-scanner effects. In contrast, SIMON—being a single-subject longitudinal dataset—showed lower volume deviations across repeated scans. Dice and Surface Dice scores were uniformly higher in SIMON, indicating improved overlap and surface-level agreement. For example, mean Dice scores for the caudate and putamen reached 0.868 and 0.897 in SIMON, compared to 0.802 and 0.848 in SRPBS. HD95 distances also decreased in SIMON (e.g., 1.234\,mm for hippocampus vs. 1.830\,mm in SRPBS), highlighting reduced boundary inconsistency. These results support the utility of repeated intra-subject data for evaluating segmentation consistency.

\paragraph{Subcortical filtering based on segmentation quality.} 
To assess the impact of quality-based filtering, we evaluated the proportion of subcortical structures removed using various thresholds on Dice and Surface Dice metrics. As summarized in Table~\ref{tab:filtering_subcortical} in Appendix A, applying a strict Surface Dice threshold of 0.92 filtered out only 5\% of regions, while retaining a low mean absolute percentage error (MAPE) across the remaining structures (2.8\% at 75th percentile, 8.6\% at 95th). Relaxing the threshold to 0.90 slightly reduced filtering (3.8\%) without degrading MAPE. In contrast, filtering with a traditional Dice threshold of 0.80 excluded more than half of all structures (52.8\%), yet retained comparable or worse error profiles. This supports the use of Surface Dice as a more efficient and precise filtering criterion for detecting outliers in automated segmentation pipelines.

       \begin{table*}[h]
        \centering
        \caption{Percentage of subcortical regions filtered out using Dice and Surface Dice thresholds, with 75th and 95th percentile MAPE values across retained regions.}
        \label{tab:filtering_subcortical}
        \begin{tabular}{lccccc}
        Filtering Metric & Threshold & Structures & \% Filtered & 75th (\% MAPE) & 95th (\%) \\
        \hline
        Surface Dice & 0.92 & Subcortical & 5.0 & 2.8 & 8.6 \\
        Surface Dice & 0.90 & Subcortical & 3.8 & 2.8 & 8.8 \\
        Dice         & 0.80 & Subcortical & 52.8 & 2.2 & 5.8 \\
        \end{tabular}
        \end{table*}

\subsection{Registration}
To compare surface-based metrics, rigid-body registration was applied using ANTs~\cite{Avants2011}. We tested two interpolation strategies: \texttt{nearestNeighbor} and \\ \texttt{genericLabel}; and two reference spaces: subject-native (first session) and standard MNI atlas. Interpolation mode affected mean volume estimates by up to 1.72\%, while template choice accounted for a smaller 0.07\% deviation.

\section{Conclusion}

This study demonstrates that even state-of-the-art segmentation tools such as FastSurfer and SynthSeg remain sensitive to scanner and protocol variability, particularly in multi-site and longitudinal settings. Despite widespread use and high reported accuracy, reproducibility across sessions and scanners remains a challenge—especially for small subcortical structures such as the amygdala and pallidum.

Our test-retest analysis on the SRPBS Traveling Subject dataset revealed excellent within-scanner consistency over five consecutive days, with volume deviations below 1\%. However, cross-site sessions introduced fluctuations up to 10\%, even when using the same individual and protocol. Similarly, in the longitudinal SIMON dataset spanning 17 years, both FastSurfer and SynthSeg showed increasing volume trends over time, but differed in magnitude and stability of outputs.

Notably, SynthSeg produced consistently larger subcortical volumes than FastSurfer (e.g., left hippocampus: 5.10\,cm$^3$ vs. 4.58\,cm$^3$), and greater inter-scan variation in cortical structures. These findings emphasize the importance of harmonization strategies or quality control filters in real-world neuroimaging pipelines.

In this study, we attempted to estimate the reliability of segmentation using various distance metrics. This approach provided a comprehensive assessment of segmentation performance beyond traditional evaluation methods. Our findings highlight the importance of employing multiple metrics to capture different aspects of segmentation quality.

While recent research focuses on speed and automation, robustness remains a bottleneck. We hope this lightweight, fully reproducible evaluation encourages more transparent benchmarking of segmentation tools on longitudinal and multi-scanner datasets.

\subsection{Work Limitations}

\subsubsection{Post-Segmentation Registration}
This study registered segmentation maps after prediction to enable geometric comparisons. However, the choice of template and interpolation method can meaningfully influence surface metrics. For instance, switching from MNI to subject-native space changed average volumes by 0.07\%, while using a non-label-preserving interpolator led to up to 1.72\% error. Future work should investigate registration-before-segmentation pipelines for robust evaluation.

\subsubsection{Lack of Preprocessing and Augmentation}
We processed raw data without denoising, intensity normalization, or augmentation to isolate the effect of domain shift. Although this reflects practical variability, it limits reproducibility.

It has been shown that classical preprocessing techniques, such as intensity normalization and histogram matching, do not consistently improve brain tumor segmentation performance across different domains. This limitation underscores the challenges posed by domain shifts in medical imaging. However, recent advancements in generative methods, including those utilizing generative adversarial networks (GANs), offer promising avenues to address these challenges. For instance, methods like M-GenSeg employ semi-supervised generative training strategies for cross-modality tumor segmentation, demonstrating improved generalization across diverse imaging modalities \cite{alefsende2022m}. Additionally, approaches that integrate GANs for synthesizing multi-modal images have been explored to enhance training data diversity and robustness \cite{gan2023multi}.

\subsubsection{No Ground Truth Labels}
Both SRPBS and SIMON datasets lack manual annotations, preventing true accuracy assessment. We evaluated reproducibility under the assumption that anatomical structures remain stable in healthy subjects, but future work should include expert-labeled benchmarks for validation.

\section*{CRediT authorship contribution statement}
\noindent
\textbf{Ekaterina Kondrateva}: Conceptualization, Methodology, Software, Formal analysis, Investigation, Writing – original draft, Writing – review \& editing. \\
\textbf{Sandzhi Barg}: Software, Data curation, Validation, Visualization, Writing – original draft, Writing – review \& editing. \\
\textbf{Mikhail Vasiliev}: Supervision, Visualization, Writing – review \& editing.

% \section{Revision}
% 		We provide, in \verb|melba.sty| a helpful command to color modifications after a revision: \verb|\revision{}|. It is automatically de-activated for papers compiled with the \verb|accepted|, \verb|arxiv| or \verb|specialissue| options.

% 	\revision{
% 		\subsection{It can also color whole sections and paragraphs}
% 			Adipisicing laborum in officia veniam in officia dolor reprehenderit ut ea sed ea reprehenderit veniam veniam culpa commodo velit commodo cillum laborum magna esse duis laboris esse in esse laborum consequat esse cupidatat.

% 			Lorem ipsum exercitation voluptate adipisicing esse cupidatat sint do excepteur laboris nisi anim mollit ut adipisicing velit quis sunt minim ut deserunt pariatur id amet elit consectetur incididunt occaecat ad labore sit in magna eiusmod.
% 	}

%%%%%%%%%%%%%%%%%%%%%%%%%%%%%%%%%%%%%%%%%%%%%%%%%%%%%%%%%%%%%%%%%%%%%%%
% Mandatory Sections. Please complete, especially for final publication
%%%%%%%%%%%%%%%%%%%%%%%%%%%%%%%%%%%%%%%%%%%%%%%%%%%%%%%%%%%%%%%%%%%%%%%

% Acknowledgements.
% Please include any funding, intellectual contributions not included in the authorship, and any other acknowledgements.
\acks{This work was supported by our genuine enthusiasm and curiosity and our hope for a brighter future of objective mental health diagnosis.
}

\section*{Declaration of Generative AI and AI-assisted technologies in the writing process}
During the preparation of this work, the authors used ChatGPT-4 to improve the readability of this paper. After using this tool, the authors reviewed and edited the content as needed and take full responsibility for the content of the publication.

% Ethical Standards.
% Please edit with the appropriate ethics considerations for your work. Include any pertinent IRB information, etc.
%
% Please note that the submission requirements included:
% The work presented must follow appropriate ethical standards in conducting research and writing the manuscript, following all applicable laws and regulations regarding treatment of animals or human subjects.
\ethics{The work follows appropriate ethical standards in conducting research and writing the manuscript, following all applicable laws and regulations regarding treatment of animals or human subjects.}

% Conflict of Interest
% Declaration of possible conflicts of interest: Authors must disclose any financial, organisational, commercial or personal conflicts of interest that might bias their work.
% If no conflicts, please say "We declare we don't have conflicts of interest."
\coi{The conflicts of interest have not been entered yet.}

% Data availability
\data{Only publicly available datasets were used.}

\bibliography{sample}

% needed in general before appendices.
% \newpage

% Appendix is optional
\clearpage
\appendix

\section{ROI descripton}

% Several software packages are widely utilized for brain morphometry analyses, each with specific strengths tailored to different research needs. The table below summarizes key features and typical applications of these tools.

% \begin{table}[h]
% \centering
% \begin{tabular}{|p{3cm}|p{7cm}|p{4cm}|}
% \hline
% \textbf{Software} & \textbf{Description} & \textbf{Citation Example} \\
% \hline
% \textbf{SPM (Statistical Parametric Mapping)} & Predominantly used for voxel-based morphometry (VBM). Operates within MATLAB and facilitates gray matter segmentation and spatial normalization. & \cite{whitwell2009voxel} \\
% \hline
% \textbf{FreeSurfer} & Specializes in surface-based morphometry and subfield segmentation. Commonly employed for hippocampal and cortical thickness measurements. & \cite{fischl2012freesurfer} \\
% \hline
% \textbf{FSL (FMRIB Software Library)} & Utilized for both VBM and diffusion imaging analyses. While less prevalent than SPM or FreeSurfer, it is often integrated into comprehensive pipeline tools. & \cite{keller2010voxel} \\
% \hline
% \end{tabular}
% \caption{Comparison of commonly used neuroimaging software for morphometric analysis.}
% \label{tab:software_comparison}
% \end{table}

\begin{table}[h]
\centering
\caption{List of evaluated ROIs: 9 bilateral cortical and 8 bilateral subcortical regions. FreeSurfer segmentation IDs are provided, cortical regions go for DKT atlas parcellation (Left, Right).}
\label{tab:roi_list}
\begin{tabular}{ll}
\textbf{Region} & \textbf{FreeSurfer IDs}
\\
\hline
Entorhinal Cortex & 1006, 2006 \\
Caudal Anterior Cingulate Cortex & 1002, 2002 \\
Inferior Parietal Cortex & 1008, 2008 \\
Fusiform Gyrus & 1007, 2007 \\
Medial Orbitofrontal Cortex & 1014, 2014 \\
Lateral Orbitofrontal Cortex & 1012, 2012 \\
Superior Temporal Cortex & 1030, 2030 \\
Insula & 1035, 2035 \\
Superior Frontal Cortex & 1028, 2028 \\
\hline
Hippocampus & 17, 53 \\
Amygdala & 18, 54 \\
Thalamus & 10, 49 \\
Caudate & 11, 50 \\
Putamen & 12, 51 \\
Pallidum & 13, 52 \\
Accumbens & 26, 58 \\
Ventral Diencephalon (VentralDC) & 28, 60 \\
\end{tabular}
\end{table}

\end{document}